\documentclass[conference]{IEEEtran}
\IEEEoverridecommandlockouts

\usepackage{cite}
\usepackage{amsmath,amssymb,amsfonts}
\usepackage{algorithmic}
\usepackage{graphicx}
\usepackage{textcomp}
\usepackage{xcolor}
\usepackage{hyperref} 
\usepackage{CJKutf8}
\usepackage{multirow} 
\usepackage{booktabs}
\usepackage{enumitem}

\def\BibTeX{{\rm B\kern-.05em{\sc i\kern-.025em b}\kern-.08em
    T\kern-.1667em\lower.7ex\hbox{E}\kern-.125emX}}
\begin{document}

\title{T-SANDHI: Tone Sandhi-aware Adaptive Network with Decoupled Hybrid Injection for Low-resource Taiwanese Hokkien Speech Recognition}

\author{
    \IEEEauthorblockN{
        Hung-Yang Sung\IEEEauthorrefmark{1}, 
        Chien-Chun Wang\IEEEauthorrefmark{2}, 
        Tien-Hong Lo\IEEEauthorrefmark{1}, 
        Yu-Sheng Tsao\IEEEauthorrefmark{3}, 
        Yung-Chang Hsu\IEEEauthorrefmark{3}, 
        Berlin Chen\IEEEauthorrefmark{1}
    }
    \IEEEauthorblockA{
        \small
        \IEEEauthorrefmark{1}Department of Computer Science and Information Engineering, National Taiwan Normal University, Taiwan \\
        \IEEEauthorrefmark{2}E.SUN Financial Holding Co., Ltd., Taiwan \\
        \IEEEauthorrefmark{3}EZAI, Taiwan
    }
}

\maketitle

\begin{abstract}
In Taiwanese Hokkien automatic speech recognition (ASR), prior studies often treat tone sandhi as a major challenge under the assumption that models fail to process implicit phonological variations.
However, our experiments on Taiwanese Hokkien reveal that speech foundation models actually handle tone sandhi variations effectively, and the real performance bottleneck stems from a localized confusion between these variations and retained citation tones. 
To address this, we propose T-SANDHI to explicitly decouple surface acoustics from underlying lexical intent on top of a frozen Whisper backbone. 
Using a lexicon-guided multi-task learning structure driven by text-derived pseudo labels, our lightweight hybrid injection module integrates independent citation and sandhi phonetic streams via dynamic gating. 
Extensive evaluation on the TAT-MOE corpus and two blind test sets demonstrates that this explicit disentanglement effectively resolves tonal mapping confusion, outperforming baselines with strict parameter efficiency.
\end{abstract}

\begin{IEEEkeywords}
automatic speech recognition, low resource, Taiwanese Hokkien, tone sandhi, lexicon-guided
\end{IEEEkeywords}

\section{Introduction}

\begin{figure}[t]
\centering
\includegraphics[width=0.95\linewidth]{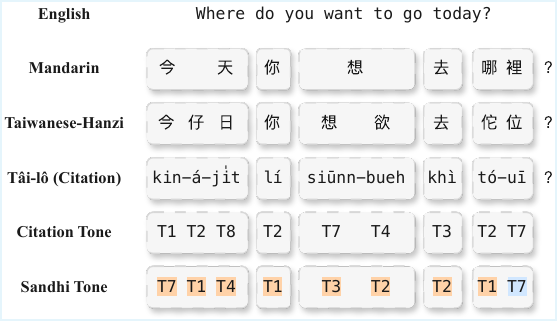}
\vspace{-10pt}
\caption{
Illustration of the acoustic-to-label discrepancy in Taiwanese Hokkien continuous speech.
Due to the tone sandhi system, the actual surface tone (sandhi tone, highlighted in orange) frequently shifts away from its dictionary pronunciation (citation tone).
For example, in the phrase ``Today where do you want to go'', the pronoun ``lí'' shifts from Tone 2 to Tone 1, whereas only the sentence-final syllable retains its original citation tone (highlighted in blue).
}
\vspace{-20pt}
\label{fig:sandhi_example}
\end{figure}

A fundamental assumption in most automatic speech recognition (ASR) systems is a consistent, reliable mapping between surface acoustics and underlying lexical units \cite{radford2023,hsu2021,graves2012,wang2021,ghodsi2020,yao2024,watanabe2017}.
However, this mapping relationship becomes less straightforward in tonal languages with complex phonological variations.
A prime example is Taiwanese Hokkien (Taiwanese) \cite{khoo2019, chen2020,chen2020a,hsieh2014,liao2022a,chou2023,lin2024}.
Unlike Mandarin \cite{zhang2022,bu2017,fu2021}, which maintains relatively static tonal mappings, Taiwanese Hokkien features a complex web of tone sandhi that applies to syllables based on their position within syntactic units \cite{cheng1968,chien2019,chou2023}.
In continuous speech, tone sandhi applies to every non-final syllable, meaning that the actual surface tone frequently shifts based on the grammatical context \cite{chou2023,chuang2025}.
Consequently, a syllable's underlying dictionary pronunciation (the \emph{citation tone}) only occurs at specific morphosyntactically defined boundaries, while all other syllables are realized with altered tones (the \emph{sandhi tone}) \cite{chuang2025,chen2023,myers2008}.
As illustrated in Fig.~\ref{fig:sandhi_example}, when a speaker utters the pronoun ``You'' (Taiwanese Hanzi\footnote{\href{https://language.moe.gov.tw/001/Upload/files/site_content/M0001/language_100/D/D005.pdf}{\nolinkurl{https://language.moe.gov.tw/.../D005.pdf}}}: \begin{CJK*}{UTF8}{bsmi}你\end{CJK*}), the citation syllable is ``lí'' (Tone 2) and the sandhi tone is Tone 1.

Current ASR systems typically rely on models to implicitly internalize these complex phonological mappings through the final transcription loss \cite{chen2023,shen2024,lin2024}.
While scaled parameters in data-abundant scenarios can partially absorb such variations, this implicit learning paradigm often faces substantial challenges under low-resource constraints \cite{bandarupa2025,klejch2025,bartelds2023,jacobs2025}.
To address these constraints, adopting parameter-efficient fine-tuning (PEFT) techniques, such as AdaLoRA, has emerged as a standard and highly effective approach to adapt speech foundation models to various low-resource languages \cite{zhang2023,song2024a,tan2025}.
Following this established paradigm, we build our study upon a parameter-efficient framework to explore foundation model behavior.
Prior study focused on utilizing syllables with citation tones as target labels, evaluating various self-supervised learning (SSL) models with connectionist temporal classification (CTC) loss \cite{chou2023}.
Their findings indicate that tone sandhi represent a major source of character substitution errors, frequently confusing phonetically similar phones in the predictions, thereby suggesting that an independent tonal handling mechanism would be beneficial for future efforts \cite{chou2023}.
However, it remains unclear whether modern speech foundation models such as Whisper are similarly affected by tone sandhi variations.

\begin{figure*}[t]
\centering
\includegraphics[width=0.95\textwidth]{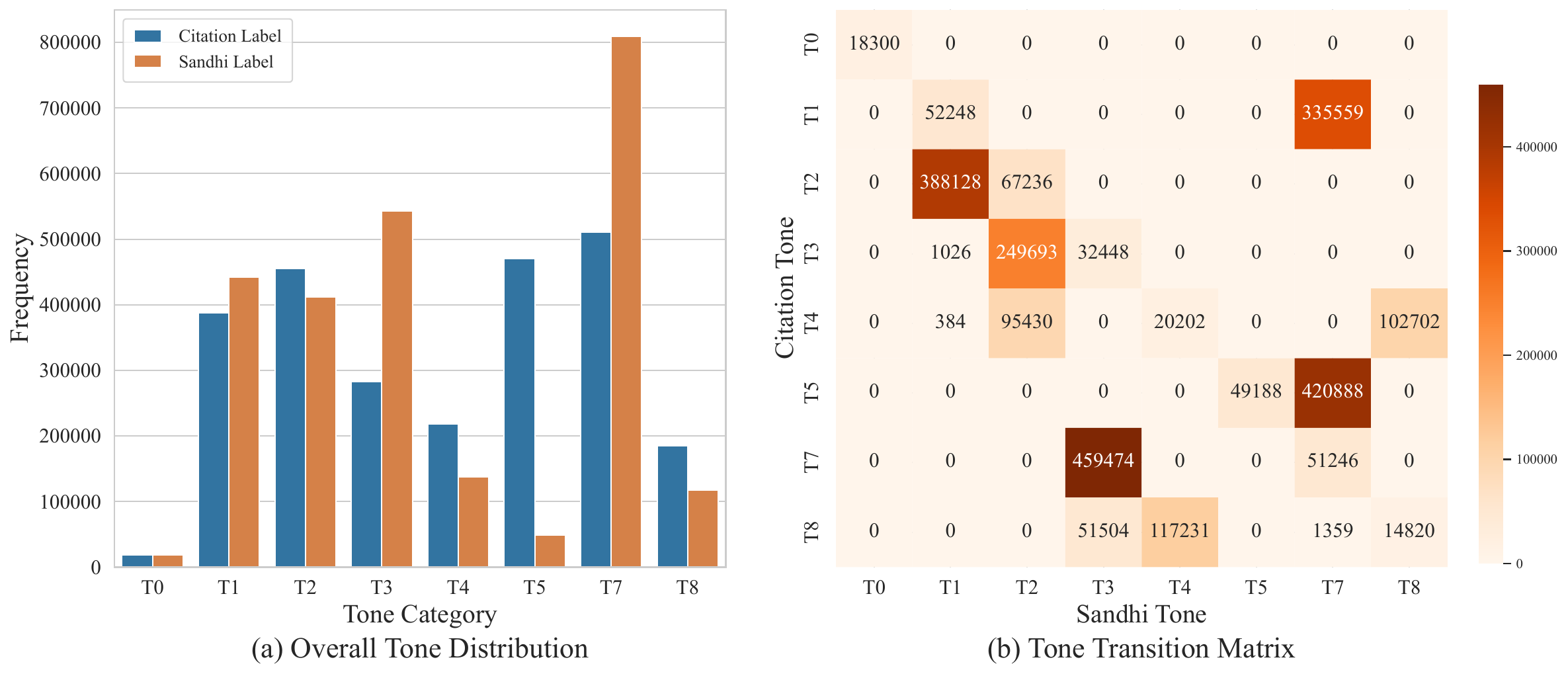}
\vspace{-10pt}
\caption{
Quantitative analysis of tonal distribution and phonetic variations across TAT-MOE corpus.
(a) Overall distribution of syllables categorized by their citation forms (blue bars) and sandhi forms (orange bars).
(b) Tone transition matrix detailing the absolute counts of syllables mapping from their citation tone to their realized tone.
}
\vspace{-15pt}
\label{fig:tone_analysis}
\end{figure*}

To address this gap, we conduct an empirical analysis to evaluate baseline predictions across various model scales.
Our findings reveal a counter-intuitive phenomenon: while modern foundation models process tone sandhi variations with reasonable proficiency, their performance bottlenecks significantly on syllables that genuinely retain their citation tones.
As we will detail in Section \ref{sec:dataset_analysis}, this localized performance gap is primarily driven by an extreme volume imbalance in continuous speech, where distinct underlying citation categories frequently conflate into identical surface realizations.
These empirical findings indicate that explicitly incorporating both citation and sandhi syllable information can serve as a viable path to reduce this localized performance gap and enhance overall speech recognition accuracy.

\begin{table}[t]
\centering
\small
\vspace{-5pt}
\caption{
Statistics of the TAT-MOE dataset across training, development and test splits, including the number of speakers, utterances and total duration in hours.
}
\vspace{-5pt}
\label{tab:tatmoe}
\setlength{\tabcolsep}{5pt}
\begin{tabular}{lcccc}
\toprule
\textbf{Split} & \textbf{\# Speakers} & \textbf{\# Utterances} & \textbf{Duration (Hours)} \\
\midrule  
Training & 328 & 86,072 & 153.33 \\
Development & 58 & 16,357 & 28.60  \\
Test & 54 & 15,962 & 26.28  \\
\midrule
Total & 440 & 118,391 & 208.21 \\
\bottomrule
\end{tabular}
\vspace{-20pt}
\end{table}

To bridge this critical gap, we propose T-SANDHI, a \textbf{T}one \textbf{S}andhi-aware \textbf{A}daptive \textbf{N}etwork designed specifically to resolve tonal mapping ambiguity.
Instead of treating the acoustic-to-label mapping as a monolithic black box, we introduce a parameter-efficient \textbf{D}ecoupled \textbf{H}ybrid \textbf{I}njection mechanism on top of a frozen Whisper backbone \cite{radford2023,song2024a,liu2024}.
To overcome data scarcity without reliance on expensive phonetic annotations, we adopt a rule-derived multi-task learning framework driven by automatically generated text-based pseudo labels.
Specifically, by leveraging CTC \cite{graves2006, kim2017, hojo2024, han2024, kusunoki2024}, we explicitly construct two auxiliary streams, where one predicts the citation syllables and the other tracks the sandhi tones.
Since sandhi mutations are highly context-dependent, a dynamic gating mechanism is employed to seamlessly integrate these dual phonetic streams.
This explicit decoupling allows the decoder to ground its predictions simultaneously on surface acoustic variations and underlying lexical intent.

The main contributions of this study are as follows:
\begin{enumerate}[noitemsep,leftmargin=*]
\item \textbf{Novel Insights on Tone Sandhi:} 
We reveal that while foundation models effectively process tone sandhi, they struggle to map these acoustics back to citation forms due to severe tonal confusion.
\item \textbf{Explicit Decoupling Architecture:} 
We propose the first ASR framework to disentangle surface acoustics from underlying lexical intent via a dynamically gated, dual-stream injection module.
\item \textbf{High Efficacy with Minimal Overhead:} 
Our approach effectively mitigates mapping ambiguity on TAT-MOE \cite{liao2022}, achieving robust performance gains while adding merely 3\% to the parameter count of the frozen backbone.
\end{enumerate}

\begin{table}[t]
\centering
\vspace{-5pt}
\caption{Baseline performance analysis on the TAT-MOE corpus, evaluated across sandhi and citation contexts.}
\vspace{-5pt}
\label{tab:preliminary_baseline}
\setlength{\tabcolsep}{0.5pt} 
\begin{tabular}{llcccc}
\toprule
\textbf{Size} & \textbf{Type (Dist.\%)} & \textbf{CER (\%)} & \textbf{Precision (\%)} & \textbf{Recall (\%)} & \textbf{F1-Score (\%)} \\
\midrule
\multirow{2}{*}{Small}  & Sandhi (87.36)   & 15.27 & 85.61 & 84.73 & 85.17 \\
                        & Citation (12.64) & 17.21 & 84.25 & 82.79 & 83.51 \\
\midrule
\multirow{2}{*}{Medium} & Sandhi (87.36)   & 13.39 & 87.32 & 86.61 & 86.96 \\
                        & Citation (12.64) & 15.21 & 85.94 & 84.79 & 85.36 \\
\midrule
\multirow{2}{*}{Large}  & Sandhi (87.36)   & 12.75 & 87.83 & 87.25 & 87.54 \\
                        & Citation (12.64) & 14.19 & 86.85 & 85.81 & 86.33 \\
\bottomrule
\end{tabular}
\vspace{-15pt}
\end{table}

\section{Corpora and Phonological Analysis}
\label{sec:dataset_analysis}

\subsection{Corpora}

To evaluate under authentic low-resource conditions, we utilized the TAT-MOE corpus \cite{liao2022}, alongside two external blind test sets: the FSRC 2020 corpus \cite{liao2020} and the yttd\_taigi\_trs corpus \cite{chen2020a}.
All three datasets feature diverse accents and spontaneous speech.
The detailed statistical partitions of the TAT-MOE corpus across the training, development, and test splits are summarized in Table \ref{tab:tatmoe}.
Acoustic signals were uniformly resampled to 16 kHz for backbone alignment.
As a tonal language, Taiwanese Hokkien exhibits a rich and complex tone sandhi system \cite{chen2023,chuang2025}.
In continuous spoken streams, these sandhi mutations apply systematically to virtually every non-final syllable within a morphosyntactically defined unit \cite{chuang2025}, resulting in a prominent discrepancy between citation dictionary forms and contextual surface realizations.
A statistical phonological analysis regarding this tonal distribution and its impact on modern ASR is detailed below.
 
\subsection{Quantitative Analysis of Tone Sandhi Impact}

To investigate the impact of this phonological discrepancy on modern speech foundation models, we evaluate Whisper baselines across various scales based on phonological boundaries.
The contextual tone sandhi variants within the training split are deterministically derived from standard tone sandhi rules \cite{chuang2025}.
As detailed in Table \ref{tab:preliminary_baseline}, scaling up the architecture improves overall performance, yet a localized gap persists.
The models consistently exhibit lower character error rates (CERs) when decoding sandhi-form syllables compared to citation-form syllables.
Regardless of model size, the corresponding drops in precision and recall confirm a tonal mapping confusion, where the networks struggle to distinguish citation features from tone sandhi variations.

To analyze this localized performance gap, we examine the data distribution from a phonological perspective.
Fig.~\ref{fig:tone_analysis}(a) highlights a severe volume imbalance in continuous Taiwanese Hokkien speech, where approximately 87\% of syllables undergo tone sandhi and only about 12\% retain their citation tones.
This extreme imbalance biases foundation models toward sandhi acoustic characteristics during pre-training.
Consequently, implicit end-to-end learning fails to construct sufficiently robust representations for citation-retained syllables under low-resource fine-tuning.
Furthermore, the tone transition matrix in Fig.~\ref{fig:tone_analysis}(b) illustrates that distinct underlying citation categories frequently conflate into identical surface realizations.
For instance, underlying Tone 1 (T1) and Tone 5 (T5) often converge into surface Tone 7 (T7) after sandhi mutations.
This many-to-one mapping ambiguity creates localized confusion between sandhi and citation forms, limiting overall performance and motivating the explicit phonetic disentanglement in our proposed architecture.

\section{Proposed Method}

\subsection{Architecture Overview}

\begin{figure}[t]
\centering
\vspace{-40pt}
\includegraphics[width=0.95\linewidth]{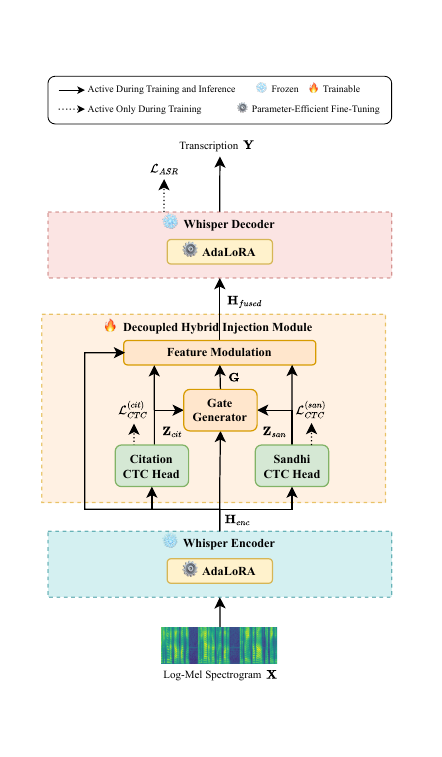}
\vspace{-45pt}
\caption{
Architecture of T-SANDHI.
To explicitly decouple lexical intent from surface acoustics, a decoupled hybrid injection module augments the frozen Whisper encoder.
It projects features into independent citation and sandhi streams, dynamically fusing them into $\mathbf{H}_{fused}$ via a gate generator.
Lightweight linear heads provide CTC supervision during training and supply decoupled representations during inference with negligible overhead.
}
\vspace{-15pt}
\label{fig:main}
\end{figure}

As established, forcing an ASR model to implicitly memorize the complex, non-linear mapping between speech undergo tone sandhi and citation labels leads to severe surface-to-underlying mapping discrepancy. 
To explicitly break this entanglement, we propose T-SANDHI. 
As illustrated in Fig.~\ref{fig:main}, the architecture builds upon the robust acoustic priors of a frozen Whisper encoder-decoder backbone. 
To maintain parameter efficiency while adapting to the intricate Taiwanese phonology, we avoid catastrophic forgetting by applying AdaLoRA \cite{zhang2023} only to the attention and feed-forward modules. 
Crucially, rather than relying on the decoder to implicitly resolve tonal ambiguities, we introduce a \emph{Decoupled Hybrid Injection} module directly on top of the encoder.
This module explicitly constructs two independent auxiliary streams: one modeling the underlying lexical intent (citation syllables), and the other tracking the actual surface tone (sandhi tones).

\subsection{Decoupled Hybrid Injection}

The core design philosophy behind our injection module is to enforce representation disentanglement without introducing heavy computational overhead. 
Let $\mathbf{H}_{enc} \in \mathbb{R}^{T \times d}$ denote the acoustic hidden states from the Whisper encoder, where $T$ is the sequence length and $d$ is the hidden dimension. 
We project these states through two auxiliary CTC heads to obtain the citation logits $\mathbf{Z}_{cit} \in \mathbb{R}^{T \times |V_{cit}|}$ and the sandhi logits $\mathbf{Z}_{san} \in \mathbb{R}^{T \times |V_{san}|}$:
\begin{align}
\mathbf{Z}_{cit} &= \mathbf{H}_{enc}\mathbf{W}_{cit} + \mathbf{b}_{cit}, \\
\mathbf{Z}_{san} &= \mathbf{H}_{enc}\mathbf{W}_{san} + \mathbf{b}_{san},
\end{align}
where $\mathbf{W}$ and $\mathbf{b}$ denote the learnable weight matrices and bias vectors, while $|V_{cit}|$ and $|V_{san}|$ represent the vocabulary sizes of the citation syllables and sandhi tones, respectively.
Crucially, we deliberately restrict these heads to simple linear projections.
This architectural bottleneck ensures that the parameter overhead remains minimal, actively forcing the core Whisper encoder to learn highly disentangled and robust acoustic representations rather than outsourcing the task to deep, heavy sub-networks.

To prepare these decoupled streams for integration, we first convert the raw logits into probability distributions via a softmax function, and then project them back to the hidden dimension $d$ to form phonetic embeddings:
\begin{align}
\mathbf{E}_{cit} &= \operatorname{Proj}_{cit}(\operatorname{Softmax}(\mathbf{Z}_{cit})), \\
\mathbf{E}_{san} &= \operatorname{Proj}_{san}(\operatorname{Softmax}(\mathbf{Z}_{san})).
\end{align}
Applying softmax grounds the projections in discrete phonetic probabilities, preventing a linear collapse of the acoustic states.

Subsequently, we employ a dynamic gating mechanism to effectively fuse these features back into the main network.
Rather than simply adding the features, we concatenate the original acoustic states with the projected phonetic embeddings to form a joint representation $\mathbf{C} = [\mathbf{H}_{enc}; \mathbf{E}_{cit}; \mathbf{E}_{san}] \in \mathbb{R}^{T \times 3d}$.
A gate generator then processes this joint representation to output a frame-level weight matrix $\mathbf{G} \in \mathbb{R}^{T \times 2}$:
\begin{equation}
\mathbf{G} = \sigma(\mathbf{C}\mathbf{W}_{gate} + \mathbf{b}_{gate}),
\end{equation}
where $\mathbf{W}_{gate} \in \mathbb{R}^{3d \times 2}$ and $\mathbf{G} = [\mathbf{g}_{cit}, \mathbf{g}_{san}]$.
The physical significance of this dynamic gate directly mirrors the context-dependent nature of Taiwanese tone sandhi.
Since tonal shifts occur strictly based on grammatical positions, $\mathbf{G}$ acts as an adaptive soft switch.
By observing both the acoustic context and the explicit phonetic hypotheses, the gate dynamically allocates attention between the underlying intent ($\mathbf{g}_{cit}$) and the surface realization ($\mathbf{g}_{san}$) frame by frame.

To prevent the newly initialized gate from catastrophically interfering with the frozen Whisper backbone during the crucial early stages of fine-tuning, we initialize the bias $\mathbf{b}_{gate}$ to a strong negative scalar ($-3$).
This insight-driven initialization forces the initial gate values toward zero, compelling the model to rely on the robust pre-trained features first and gradually learn to blend in the decoupled phonetic patches.
The final fused representation $\mathbf{H}_{fused}$, which is subsequently fed to the decoder, is computed as:
\begin{equation}
\mathbf{H}_{fused} = \mathbf{H}_{enc} + \mathbf{g}_{cit} \odot \mathbf{E}_{cit} + \mathbf{g}_{san} \odot \mathbf{E}_{san},
\end{equation}
where $\odot$ represents element-wise multiplication, with $\mathbf{g}_{cit}$ and $\mathbf{g}_{san}$ implicitly broadcasted across the hidden dimension $d$.

\subsection{Multi-Task Learning Objective}

To actualize this decoupled architecture, the training objective must explicitly penalize entangled representations. 
Our primary objective remains the standard sequence-to-sequence cross-entropy loss $\mathcal{L}_{ASR}$ generated by the Whisper decoder for the final Taiwanese Hanzi transcriptions. 
However, to provide the necessary phonetic grounding for our injection module, we must introduce auxiliary losses. 
Since frame-level forced alignment data is prohibitively expensive and largely unavailable for low-resource languages, we formulate these as CTC objectives.
CTC naturally marginalizes over all possible unsegmented acoustic alignments. 
Thus, $\mathcal{L}_{CTC}^{(cit)}$ explicitly guides the extraction of citation syllables, while $\mathcal{L}_{CTC}^{(san)}$ supervises the sandhi tone tracking:
\begin{equation}
\mathcal{L}_{Total} = \mathcal{L}_{ASR} + \lambda_{cit} \mathcal{L}_{CTC}^{(cit)} + \lambda_{san} \mathcal{L}_{CTC}^{(san)},
\label{eq:total}
\end{equation}
where $\lambda_{cit}$ and $\lambda_{san}$ are scalar hyperparameters. 
This joint optimization ensures that the encoder effectively disentangles ``what is heard'' from ``what is meant'' before passing the representation to the decoder.

\section{Experimental Setup}

\subsection{Dual-Track Supervision Strategy}

Our data preparation goes beyond formatting to construct explicit supervision signals for the decoupled architecture.
Using the official MOE Taiwanese Dictionary\footnote{\url{https://sutian.moe.edu.tw/zh-hant}} and Taibun toolkit\footnote{\url{https://github.com/andreihar/taibun}}, we extracted two strictly parallel phonetic tracks from the Hanzi transcriptions: the citation dictionary forms and the contextual sandhi tones.
This dual-track extraction yields the text-derived pseudo labels to independently guide our auxiliary CTC streams under weak supervision.
Specifically, the citation vocabulary $|V_{cit}|$ comprises 1309 unique syllables with tone marks, while the sandhi vocabulary $|V_{san}|$ tracks 10 classes (8 tones, a neutral tone, and a blank token).

\subsection{Implementation Details}

We employed Whisper-small \cite{radford2023} as our acoustic foundation.
To preserve its pre-trained speech representations while adapting to Taiwanese Hokkien, we did not perform full fine-tuning.
Instead, we applied AdaLoRA \cite{zhang2023} to the attention and feed-forward blocks with an initial rank of 12 (pruned down to 4) and a dropout rate of 0.1 to ensure parameter efficiency.
To guarantee stable training for our decoupled hybrid injection module, the bias vector of the gate mechanism, $\mathbf{b}_{\text{gate}}$, was initialized to $-3$.
This configuration ensures that the model heavily relies on the robust frozen backbone during the initial training phase, gradually incorporating the auxiliary phonetic information as training converges.
For the multi-task objective function (Eq. \ref{eq:total}), the CTC loss weights $\lambda_{\text{cit}}$ and $\lambda_{\text{san}}$ were empirically set to 0.9 and 0.1, respectively.
This weighting ensures adequate phonetic supervision without overshadowing the primary sequence-to-sequence objective.

\begin{table*}[t]
\small
\centering
\caption{
CER (\%) comparison on the TAT-MOE, FSRC 2020, and yttd\_taigi\_trs blind test sets.
}
\vspace{-5pt}
\setlength{\tabcolsep}{13pt}
\begin{tabular}{lccccc}
\toprule
\multirow{2}{*}{\bf Model} & \multirow{2}{*}{\bf Parameters (M)} & \multicolumn{2}{c}{\bf TAT-MOE} & \bf FSRC 2020 & \bf yttd\_taigi\_trs \\
\cmidrule(lr){3-4}
\cmidrule(lr){5-5}
\cmidrule(lr){6-6}
 & & \bf Development & \bf Test & \bf Blind Test & \bf Blind Test \\
\toprule
Zipformer \cite{yao2024}         & 65   & 48.57 & 45.82 & 15.69 & 58.15 \\
HuBERT-base \cite{hsu2021}       & 96   & 26.16 & 24.49 & 12.97 & 61.68 \\
CLiFT-ASR \cite{sung2025}        & 96   & 22.37 & 20.94 &  8.06 & 57.69 \\
\midrule
Whisper-small (Encoder-CTC)      & 104  & 19.90 & 17.98 &  9.43 & 42.73 \\
Whisper-small (Fully Fine-tuned) & 244  & 22.47 & 18.68 &  7.66 & 44.43 \\
Whisper-small (AdaLoRA)          & 244  & 18.54 & 16.30 &  8.75 & 39.11 \\
\bf T-SANDHI (Ours)              & \bf 249 & \bf 16.49 & \bf 14.30 & \bf 7.36 & \bf 36.57 \\
\midrule
Qwen3-ASR                        & 600  & 16.74 & 14.03 &  7.38 & 44.91 \\
Whisper-large (AdaLoRA)          & 1,550 & 16.03 & 13.89 &  7.20 & 29.31 \\
\bottomrule
\end{tabular}
\label{tab:main_results}
\vspace{-15pt}
\end{table*}

\subsection{Evaluation and Diagnostic Metrics}

Our primary performance metric is the CER on Taiwanese Hanzi, which directly reflects practical utility.
However, to evaluate whether our architecture effectively addresses the localized mapping confusion between surface acoustics and underlying lexical intent, we must look beyond final transcription errors.
Therefore, we introduce two diagnostic metrics: syllable error rate (SER) to evaluate underlying lexical intent tracking, and tone error rate (TER) to measure surface tone resolution.
Tracking these metrics allows us to empirically verify that the performance gains stem directly from our explicit phonetic disentanglement.

\begin{table}[t]
\centering
\small
\caption{
CER (\%) performance and relative reduction (Rel., \%) across Whisper scales on the TAT-MOE corpus.
}
\vspace{-5pt}
\label{tab:model_scale_results}
\setlength{\tabcolsep}{13pt} 
\begin{tabular}{lccc}
\toprule
\bf Model Scale & \bf Baseline & \bf T-SANDHI & \bf Rel. \\
\midrule
Small  & 16.30 & 14.30 & 12.27 \\
Medium & 14.58 & 13.65 & 6.38  \\
Large  & 13.89 & 12.93 & 6.91  \\
\bottomrule
\end{tabular}
\vspace{-15pt}
\end{table}

\section{Results and Discussion}

\subsection{Overall ASR Performance}

Table \ref{tab:main_results} demonstrates that explicitly decoupling surface contextual features from underlying canonical features effectively mitigates the tonal mapping confusion caused by tone sandhi.
To ensure a rigorous evaluation, our baselines span three architectural paradigms: RNN-Transducer (RNN-T) models, pure CTC frameworks, and attention-based encoder-decoder (AED) foundation models. 
Traditional end-to-end RNN-T models struggle to implicitly memorize dynamic tonal mappings, as demonstrated by Zipformer \cite{yao2024}, HuBERT-base \cite{hsu2021}, and CLiFT-ASR \cite{sung2025} yielding Test CERs of 45.82\%, 24.49\%, and 20.94\%, respectively.
Augmenting the foundation encoder with a standard CTC head (Whisper-small (Encoder-CTC)) improves performance but yields a suboptimal 17.98\% CER, constrained by a monolithic sequence loss.
Naive adaptations of the Whisper backbone similarly fall short: full fine-tuning suffers from representation distortion (18.68\%), and standard AdaLoRA (16.30\%) remains bottlenecked by entangled phonetic features.

Crucially, T-SANDHI overcomes these limitations.
With only a 5M parameter overhead for the decoupled hybrid injection module, our framework achieves a 14.30\% Test CER, a 12.27\% relative error reduction over AdaLoRA.
Coupled with robust out-of-domain generalization on the FSRC 2020 (7.36\%) and yttd\_taigi\_trs (36.57\%) blind test sets, these results confirm that explicit phonetic disentanglement, rather than mere parameter scaling, drives the performance gains.

\subsection{Scalability Across Backbone Capacities}

We further investigate T-SANDHI's scalability across larger foundation models.
As detailed in Table \ref{tab:model_scale_results}, the method yields consistent gains across all evaluated Whisper backbones.
While increasing model capacity naturally lowers the baseline error rate, our dual-stream module extracts additional improvements, achieving relative error reductions of 12.27\%, 6.38\%, and 6.91\% on the small, medium, and large backbones, respectively.
This confirms that phonetic disentanglement provides orthogonal benefits to naive parameter scaling.

\begin{table}[t]
\small
\caption{Ablation studies of T-SANDHI on the TAT-MOE corpus.}
\vspace{-5pt}
\label{tab:ablation}
\centering
\setlength{\tabcolsep}{7pt}
\begin{tabular}{lccc}
\toprule
\bf Model Configuration & \bf SER \% & \bf TER \%  & \bf CER \%\\
\midrule
T-SANDHI (Full Model)                  &  16.27 & 13.99 & 14.30 \\
\midrule
\quad w/o Decoupled Streams &   -    &   -   & 16.30 \\
\quad w/o Citation CTC Head &   -    & 13.52 & 15.44 \\
\quad w/o Sandhi CTC Head   & 16.68  &   -   & 15.19 \\
\midrule
\quad w/o Gate Generator    & 17.45  & 14.30 & 15.55 \\
\quad w/o Dynamic Gating    & 17.36  & 14.19 & 15.67 \\
\bottomrule
\end{tabular}
\vspace{-10pt}
\end{table}

\subsection{Ablation Studies}

Table \ref{tab:ablation} validates that our performance gains stem from explicit phonetic disentanglement.
Removing the dual-stream architecture entirely regresses the CER to 16.30\%, indicating that the frozen backbone alone cannot fully resolve acoustic-to-lexical confusion.
Ablating individual CTC heads highlights the mechanics of tone sandhi resolution.
Omitting the citation head improves the surface TER (13.52\%) but degrades the final CER (15.44\%), as the model overfits surface acoustics and loses underlying Hanzi identity.
Conversely, removing the sandhi head worsens the SER (16.68\%), demonstrating that surface tracking provides essential acoustic grounding.

Moreover, optimal integration requires dynamic fusion.
An unweighted summation degrades CER to 15.55\% by re-entangling features, and using static global weights (heuristically tuned on the development set) yields a suboptimal 15.67\%.
Static weights fail to capture position-dependent sandhi rules (e.g., word-medial mutation vs. sentence-final retention), underscoring the necessity of a frame-by-frame neural soft-switch.

\begin{table}[t]
\small
\centering
\caption{
Ablation study on the granularity of injected linguistic information for the dual streams.
}
\vspace{-5pt}
\label{tab:stream_granularity}
\setlength{\tabcolsep}{15pt}
\begin{tabular}{ccc}
\toprule
\bf Citation Info & \bf Sandhi Info & \bf Test CER (\%) \\
\midrule
Tone     & Tone     & 15.60 \\
Tone     & Syllable & 15.84 \\
Syllable & Tone     & \textbf{14.30} \\
Syllable & Syllable & 14.80 \\
\bottomrule
\end{tabular}
\vspace{-5pt}
\end{table}

\begin{figure}[t]
\centering
\small
\includegraphics[width=1.0\linewidth]{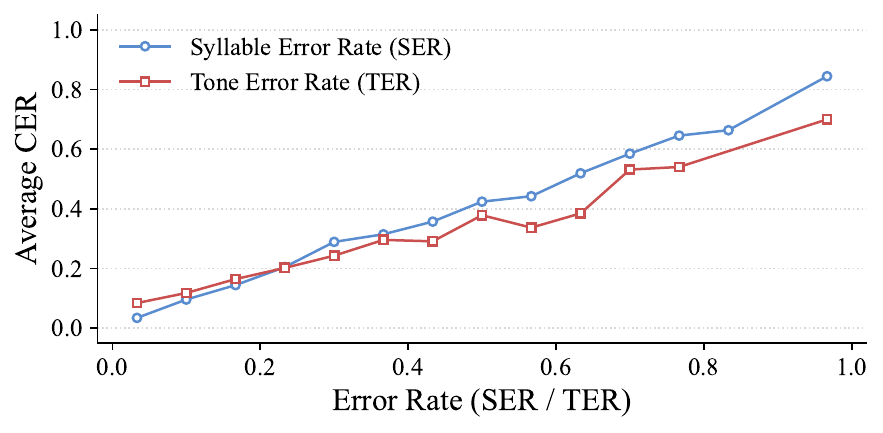}
\vspace{-20pt}
\caption{
Correlation between phonetic errors and transcription accuracy.
}
\vspace{-10pt}
\label{fig:error_trend}
\end{figure}

\subsection{Impact of Information Granularity in Dual Streams}
\label{sub:granularity_analysis}

To determine the optimal feature resolution, we ablated the linguistic granularity injected into the dual streams.
Table \ref{tab:stream_granularity} contrasts configuring either stream with phone-level tone labels versus full syllable-level targets.
Pairing citation syllables with surface tones yields the lowest CER (14.30\%), outperforming all symmetric configurations.

Phonetically, the citation stream relies on full syllables to anchor lexical identity.
Conversely, the surface stream, tracking actual acoustic realizations, functions best when constrained to tone-level variations.
This structural bottleneck prevents the sandhi head from overfitting to specific lexical items, effectively mitigating localized mapping confusion in low-resource settings.

\subsection{Correlation between Phonetic and Character Errors}

Fig.~\ref{fig:error_trend} plots final CER against auxiliary phonetic error rates, verifying that the decoder actively utilizes decoupled representations.
The strong positive correlation confirms that accurate phonetic grounding is a prerequisite for transcription.
Crucially, sandhi TER and citation SER trajectories diverge as errors increase.
The SER's steeper slope indicates the citation syllable serves as the primary structural anchor; misidentifying it severely degrades character prediction.
The TER's gentler slope suggests the decoder leverages contextual language modeling to tolerate minor tone errors.
Nevertheless, minimizing TER remains essential to resolve localized mapping confusion and push CER below 20\%.
This hierarchy, where syllables provide structure and tones provide disambiguation, empirically justifies the dual-stream strategy.

\subsection{Emergent Phonological Awareness in Dynamic Gating}

\begin{figure}[t]
\centering
\includegraphics[width=1.0\linewidth]{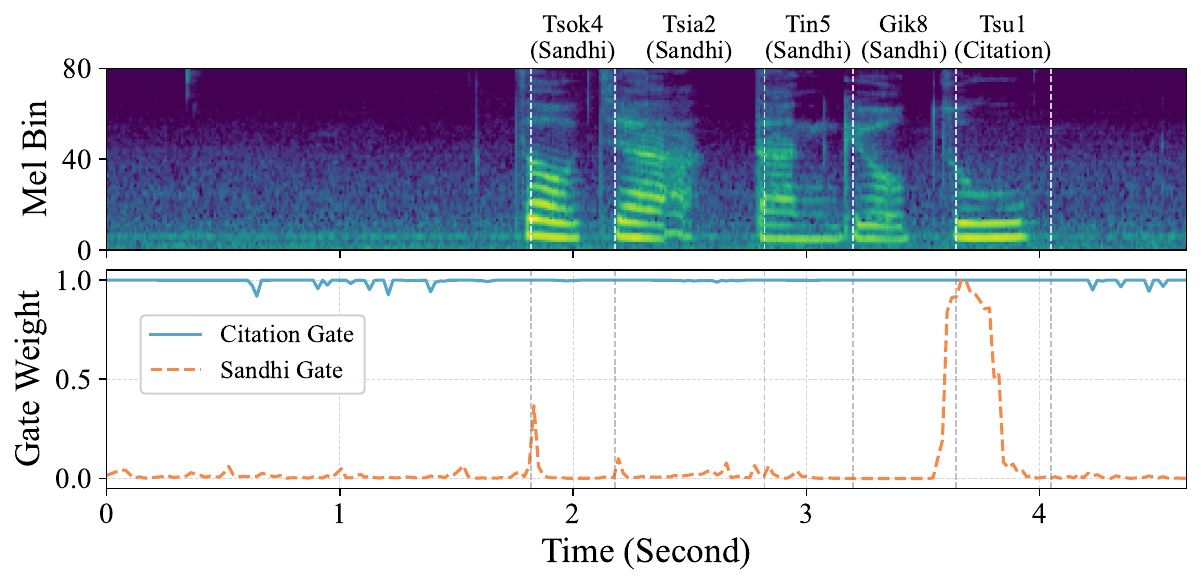}
\vspace{-20pt}
\caption{
Frame-level visualization of the dynamic gate.
}
\label{fig:gating}
\end{figure}

Fig.~\ref{fig:gating} visualizes dynamic gate behavior across a continuous utterance to illustrate mapping resolution.
During phrase-internal mutations (e.g., \textit{Tsok4} realized as \textit{Gik8}), the network autonomously suppresses the surface stream ($\mathbf{g}_{\text{san}} \approx 0$) to filter deceptive acoustics, anchoring instead on the citation intent ($\mathbf{g}_{\text{cit}}$).
At the phrase-final position (\textit{Tsu1}), where surface acoustics and citation tones align, $\mathbf{g}_{\text{san}}$ activates sharply ($\mathbf{g}_{\text{san}} \to 1$).
This demonstrates the network learns to rely on surface acoustics only when phonologically reliable.
These findings indicate an emergent, data-driven awareness of right-prominent sandhi rules within the gating mechanism.

\begin{figure}[t]
\centering
\includegraphics[width=0.7\linewidth]{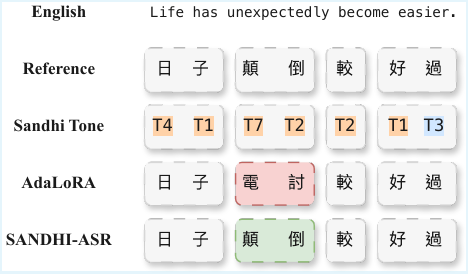}
\caption{
Qualitative comparison of error resolution.
}
\vspace{-10pt}
\label{fig:error_example}
\end{figure}

\subsection{Targeted Error Analysis}

Fig.~\ref{fig:error_example} illustrates T-SANDHI's resolution of character mapping confusion.
In the phrase ``\begin{CJK*}{UTF8}{bsmi}顛倒\end{CJK*}'' (\textit{tian-tò}, unexpectedly), underlying citation tones (T1-T3) mutate into surface tones (T7-T2).
Misled by these surface acoustics, the Whisper-small (AdaLoRA) baseline incorrectly predicts the phonetically identical but semantically unrelated ``\begin{CJK*}{UTF8}{bsmi}電討\end{CJK*}'' (\textit{tiān thó}, citation T7-T2).
This substitution highlights how conventional models, lacking explicit guidance, overfit surface tone variations and fail to recover lexical intent.
In contrast, T-SANDHI successfully outputs the correct Hanzi.
By tracking surface realizations (T7-T2) while anchoring on citation identity (T1-T3), the decoupled framework bridges the phonological gap.
This frame-by-frame dynamic gating effectively resolves the confusion of entangled architectures.

\section{Conclusion and Future Work}

In this paper, we proposed T-SANDHI\footnote{Our source code: \url{https://github.com/redsheep913/T-SANDHI.git}}, a parameter-efficient framework that explicitly disentangles surface acoustics from underlying lexical intent to mitigate tone sandhi mapping confusion.
By employing a dynamic gating mechanism, our approach achieves robust performance gains across foundation models without substantial parameter scaling.
While currently supervised by rule-derived pseudo-labels, future work will explore unsupervised disentanglement for undocumented dialects, as well as extending this decoupled paradigm to Mandarin-Taiwanese code-switching.

\newpage

\bibliographystyle{IEEEtran}
\begingroup
\bibliography{references}
\endgroup

\end{document}